\documentclass[10, times]{elsarticle}

\usepackage{amssymb}
\usepackage{amsmath}
\usepackage{graphicx}
\usepackage{orcidlink}
\usepackage{here}
\usepackage{natbib}
\usepackage{algorithm}
\usepackage{algpseudocode}
\usepackage{multirow}
\usepackage{setspace}
\DeclareUnicodeCharacter{2061}{}

\journal{Elsevier}

\begin{document}

\begin{frontmatter}



\title{Decomposing one-class support vector machine into an ensemble of one-data support vector machines} 


\author[a]{Toshitaka Hayashi \orcidlink{0000-0002-7599-4404}}
\ead{toshitaka.hayashi@uhk.cz}
\author[a]{Dalibor Cimr \orcidlink{0000-0003-2197-8553}}
\ead{dalibor.cimr@uhk.cz}
\author[a,b,c]{Hamido Fujita\corref{cor1} \orcidlink{0000-0001-5256-210X}}
\ead{fujitahamido@utm.my, HFujita-799@acm.org}
\author[a]{Richard Cimler \orcidlink{0000-0001-6712-9894}}
\ead{richard.cimler@uhk.cz}
\cortext[cor1]{Corresponding author: Professor Hamido Fujita \\ Email: fujitahamido@utm.my, HFujita-799@acm.org \\  Postal address: Kotorizawa, 2-27-5, Morioka, Iwate 020-0104, Japan\\  Tel: +81 8067208218}
\affiliation[a]{organization={Faculty of Science, University of Hradec Kralove},
            addressline={Hradecká 1285}, 
            city={Hradec Kralove},
            postcode={50003}, 
            country={Czech Republic}}
\affiliation[b]{organization={Malaysia-Japan International Institute of Technology (MJIIT), Universiti Teknologi Malaysia},
            addressline={Jalan Sultan Yahya Petra, Kampung Datuk Keramat}, 
            city={Kuala Lumpur},
            postcode={54100}, 
            country={Malaysia}}
\affiliation[c]{organization={Regional Research Center, Iwate Prefectural University},
            addressline={Sugo 152-52}, 
            city={Takizawa},
            postcode={020-0693}, 
            state={Iwate},
            country={Japan}}

\begin{abstract}
One-class Support Vector Machine(OCSVM) is a variant of support vector machine(SVM) for one-class classification (OCC). However, OCSVM has scalability issues with large datasets. This paper proposes an acceleration strategy for OCSVM. The idea is to decompose the dataset into samples and train OCSVM models for single data points. Subsequently, ensemble learning is applied to combine all models to compute the OCSVM model for the dataset. In addition, further acceleration is achieved through a support vector selection strategy with an OCSVM model trained on the average of the training samples. The experiment compared the proposal and traditional OCSVM using the Python package. The proposed strategy is faster than traditional OCSVM, while achieving similar classification results. Moreover, the proposed strategy can create a one-to-one correspondence between samples and models. Source code is uploaded at https://github.com/ToshiHayashi/ODSVM.
\end{abstract}

\begin{highlights}
\item This paper accelerates the one-class support vector machine (OCSVM).
\item The bottleneck of OCSVM is the optimization process for the large-scale dataset.
\item The main idea is to decompose the dataset into single samples.
\item One-data support vector machine (ODSVM) is an OCSVM for a single sample.
\item The ensemble of ODSVMs is faster than OCSVM and achieves similar classification results.
\end{highlights}

\begin{keyword}
One-class classification \sep One-class support vector machine \sep Ensemble Learning \sep Soft Computing \sep Support Vector Machine

\end{keyword}

\end{frontmatter}
\section{Introduction}
\label{sec1}
One-class classification (OCC) is a supervised classification problem where the training dataset contains only one class. The goal is to classify data into one known class and the remaining unknown classes. OCC is the simplest problem setting to detect an unknown class. 

One-class support vector machine (OCSVM) \cite{[1]} is an OCC algorithm extended from the support vector machine (SVM). The original SVM \cite{[20]} is a binary classification algorithm to maximize the margin between support vectors (data points near the boundary). The core idea behind SVM is that only support vectors matter for classification, while the remaining samples are unimportant. OCSVM applies the same idea to a single class. The support vectors are samples around the boundary between one class and the remaining unknown classes. In other words, support vectors are the most abnormal training samples belonging to one class. 

OCSVM \cite{[1]} is a widely used OCC algorithm for feature vectors. Several comparisons \cite{[11],[35]} suggest that OCSVM is the top-level OCC algorithm for feature vectors. Many researchers proposed variants of OCSVMs to improve accuracy metrics \cite{[18],[19]} and accelerate computation \cite{[4],[15]}. However, OCSVM has scalability issues for large datasets \cite{[4],[21]}. This aspect is a challenge for handling big data. 

The motivation of this paper is to accelerate the OCSVM algorithm. The main bottleneck is the time complexity on a large dataset, where most samples are not support vectors. These unimportant samples increase dataset size and computation cost. 

To remove this bottleneck, this study decomposes the dataset into single samples and trains OCSVM models for each sample. In other words, this paper decomposes OCSVM into an ensemble of one-data support vector machine (ODSVM) models. ODSVM is an OCSVM trained on a single sample. Unlike the standard OCSVM, ODSVM does not solve the optimization problem because the training data is a single point. Moreover, further acceleration is achieved by a faster support vector selection strategy using an ODSVM trained on the average of the training samples. Accordingly, the proposed method reduces the time complexity.

On the other hand, ensemble learning has four alternatives: 1) Bagging \cite{[29]} combines the models from the same algorithm, trained on different datasets. 2) Voting \cite{[30]} combines the models from different algorithms, trained on the same dataset. 3) Stacking \cite{[31]} utilizes the outputs of base learners as feature vectors of the ensemble model. 4) Boosting \cite{[32]} connects the models sequentially, where the first model learns samples, while the remaining models learn from the errors of previous models. This paper applies bagging because it is the fastest alternative. Boosting and stacking are also applicable. However, they are more time-consuming than bagging. In addition, applying voting is conceptually impossible, as all ODSVM models learn different samples.

This paper has the following contributions:
\begin{itemize}
\item The originality is to decompose OCSVM into an ensemble of ODSVMs trained on single data points.
\item  This paper claims that support vectors in OCSVM are the most abnormal training samples, and optimization is unnecessary to compute these samples.
\item The ensemble of ODSVMs is faster than OCSVM, while keeping similar AUC scores. This advantage could be extended to original SVM.
\item ODSVM can establish a one-to-one correspondence between samples and models. This feature is theoretically useful because one can treat samples as models or vice versa. 
\end{itemize}

The paper is organized as follows. Section \ref{sec2} describes related work. Section \ref{sec3} decomposes OCSVM into ODSVMs. Sections \ref{sec4} and \ref{sec5} present the experimental results and the discussion, respectively. Finally, Section \ref{sec6} is the conclusion. 

\section{Related Work}
\label{sec2}
This paper is related to one-class classification and ensemble learning. Subsection \ref{sec2.1} provides a brief introduction to one-class classification and OCSVM. Subsection \ref{sec2.2} describes ensemble learning.

\subsection{One-class classification}
\label{sec2.1}
One-class classification (OCC) is a classification problem where the training dataset contains only one class. Equation (\ref{eq1}) shows the general OCC framework, which uses the score function and the threshold value $\lambda$:

\begin{equation}\label{eq1}
OCC(X)=\left\{\begin{alignedat}{2}
One (score(X)\ge\lambda) \\
Other (score(X)< \lambda)
\end{alignedat}
\right.,
\end{equation}
where X is a sample. The score function computes normality from samples. 

\begin{equation}\label{eq2}
score:X \rightarrow normality
\end{equation}
Normality refers to the likelihood that a sample belongs to a single class. The main challenge in OCC research is designing the score function. Several OCC algorithms have been proposed. They can be grouped into boundary-based methods \cite{[1],[8]}, distance-based methods \cite{[9],[11]}, probability-based methods \cite{[10]}, fake-based methods \cite{[27]}, and subtask-based methods \cite{[28]}. This paper aims to improve OCSVM, which falls into a boundary-based approach. 

\subsubsection{One-class support vector machine}
\label{sec2.1.1}
The OCSVM \cite{[1]} uses score function as shown in equation (\ref{eq3}):
\begin{equation}\label{eq3}
score(x)=\sum_{i=1}^n \alpha_i * K(x_i, x)-\rho
\end{equation}
where n is the number of training samples, $\alpha$ is the weight of sample ($\alpha_i \ge 0$, and  $\sum_{i=1}^n \alpha_i =1$), and $\rho$ is the value to handle threshold. Moreover, $x_i$ is called a support vector when $\alpha_i>0$. In addition, K refers to the kernel function: the sklearn package \cite{[6]} provides four alternatives, linear, polynomial, RBF (Radial Basis Function, also called Gaussian \cite{[11]}), and sigmoid. Equations (\ref{eq4})-(\ref{eq7}) show kernel functions:
\begin{equation}\label{eq4}
linear(x_i,x)= 〈x_i,x〉
\end{equation}

\begin{equation}\label{eq5}
polynomial(x_i,x)=(\gamma *〈x_i,x〉+ Coef_0)^{degree}
\end{equation}

\begin{equation}\label{eq6}
RBF(x_i,x)=exp(-\gamma * || x_i-x ||^2)
\end{equation}

\begin{equation}\label{eq7}
sigmoid(x_i,x)=tanh(\gamma *〈x_i,x〉+ Coef_0)
\end{equation}
where $〈x_{tr},x〉$ is the dot product. Moreover, $\gamma$, $Coef_0$, and degree are hyperparameters of kernels. The literature \cite{[11],[21]} reported that the RBF is the best kernel for OCSVM. 

The main idea of OCSVM was to map samples into kernel space and to maximize the hyperplane between support vectors and the origin O. Schölkopf \cite{[1]} defined the primal problem of OCSVM as follows:
\begin{equation}
\min_{w,\rho,\xi} \frac{1}{2}\|w\|^2 - \rho + \frac{1}{\nu n}\sum_{i=1}^{n}\xi_i
\end{equation}
subject to $w^\top x_i \ge \rho - \xi_i$, and $\xi_i \ge 0$, where n is the number of samples, $\xi_i$ is a slack variable, w is a regularization term, and $\rho$ is the threshold value. 

Subsequently, Schölkopf \cite{[1]} defined the dual problem as follows:
\begin{equation}\label{eq8}
\min_{\alpha}⁡ \frac{1}{2}\sum_{i=1}^{n}\sum_{j=1}^{n}{\alpha_i}{\alpha_j}K(x_i,x_j)
\end{equation}
where $\alpha$ is weights that $0 \le \alpha_i \le 1/vn$ and $\sum_{i=1}^n \alpha_i =1$. Moreover, $n$ is the number of samples, while $v$ (also called nu in sklearn \cite{[6]}) is a hyperparameter to control the trade-off \cite{[1]}. This paper simplifies the dual problem by assigning n = 1.

OCSVM is also known as support vector data description (SVDD) \cite{[12]}, which is a problem to compute a hypersphere around the training samples. OCSVM and SVDD are equivalent when RBF (Gaussian) kernels are used \cite{[11]}.

OCSVM has several variants. Manevitz et al. \cite{[18]} and Li et al. \cite{[19]} treated part of the training samples as outliers. These ideas were based on the hypothesis that training data includes outliers. Moreover, Hao et al. \cite{[24]} introduced a fuzzy function to OCSVM. In addition, Yang et al. \cite{[25]} proposed the neighborhood OCSVM for time series data. The idea was to treat timestamps that were similar as neighbors. Furthermore, Ruff et al.\cite{[36]} proposed a deep SVDD algorithm for an image dataset that combined deep learning and OCSVM.

Other researchers improved the processing speed of OCSVM. Least square OCSVM \cite{[15]} reduced the computation cost of OCSVM from O($n^3$) to O($n^2$) \cite{[5]}. 
Stochastic Gradient Descent validates the model using a few random samples rather than the entire dataset. Nystrom method \cite{NystromSVM} is a technique to approximate large kernel matrices with small subsets. Alam et al. \cite{[22]} proposed a sample reduction strategy by estimating the farthest boundary point. 

Other researchers separated datasets into smaller units. Krawczyk et al. \cite{[2]} applied k-means clustering to the dataset and combined OCSVM models trained on each cluster. 

Kang et al. \cite{[4]} estimated support vectors from an ensemble of multiple OCSVM models trained on a few bootstrap samples. Subsequently, OCSVM models are trained using only estimated support vectors. However, these methods require an optimization process for the small dataset. 

This paper decomposes the dataset into single samples. The proposed method eliminates the time complexity for large datasets. To the best of our knowledge, the existing OCSVM papers did not decompose the dataset into data points. 

\subsection{Ensemble Learning}
\label{sec2.2}
Ensemble learning \cite{[26]} is a technique to combine multiple machine learning models. 
As described in the introduction, ensemble learning is categorized into four types. Bagging \cite{[29]} combines models from the same algorithm trained on different datasets, while voting \cite{[30]} combines models from different algorithms trained on the same dataset. Stacking \cite{[31]} uses the model outputs as feature vectors of the ensemble model. Boosting \cite{[32]} learns the errors of models. The proposed method falls into the bagging approach because it combines multiple ODSVM models trained on different data points. 

Ensemble learning has been applied to combine SVM, OCSVM, and OCC models. For instance, Xing et al. \cite{[16]} applied AdaBoost for OCSVMs. 
Other researchers decomposed a complex problem into simple problems. One example is decomposing datasets into smaller units, such as clusters \cite{[2]} or bootstrap samples \cite{[4]}. Moreover, Shorab et al. \cite{[13]} decomposed SVDD into a subspace-level problem and considered ensemble learning. Furthermore, Krawczyk et al. \cite{[3]} decomposed a multi-class classification problem into an ensemble of OCC problems. 

Similarly, this paper decomposes the dataset into an ensemble of single data points. In other words, the proposed method decomposes OCC into an ensemble of one-data classification problems. To the best of our knowledge, this is a new research direction.

\section{The proposed method}
\label{sec3}
\begin{figure}[H]
\centering
\includegraphics[width=1\linewidth]{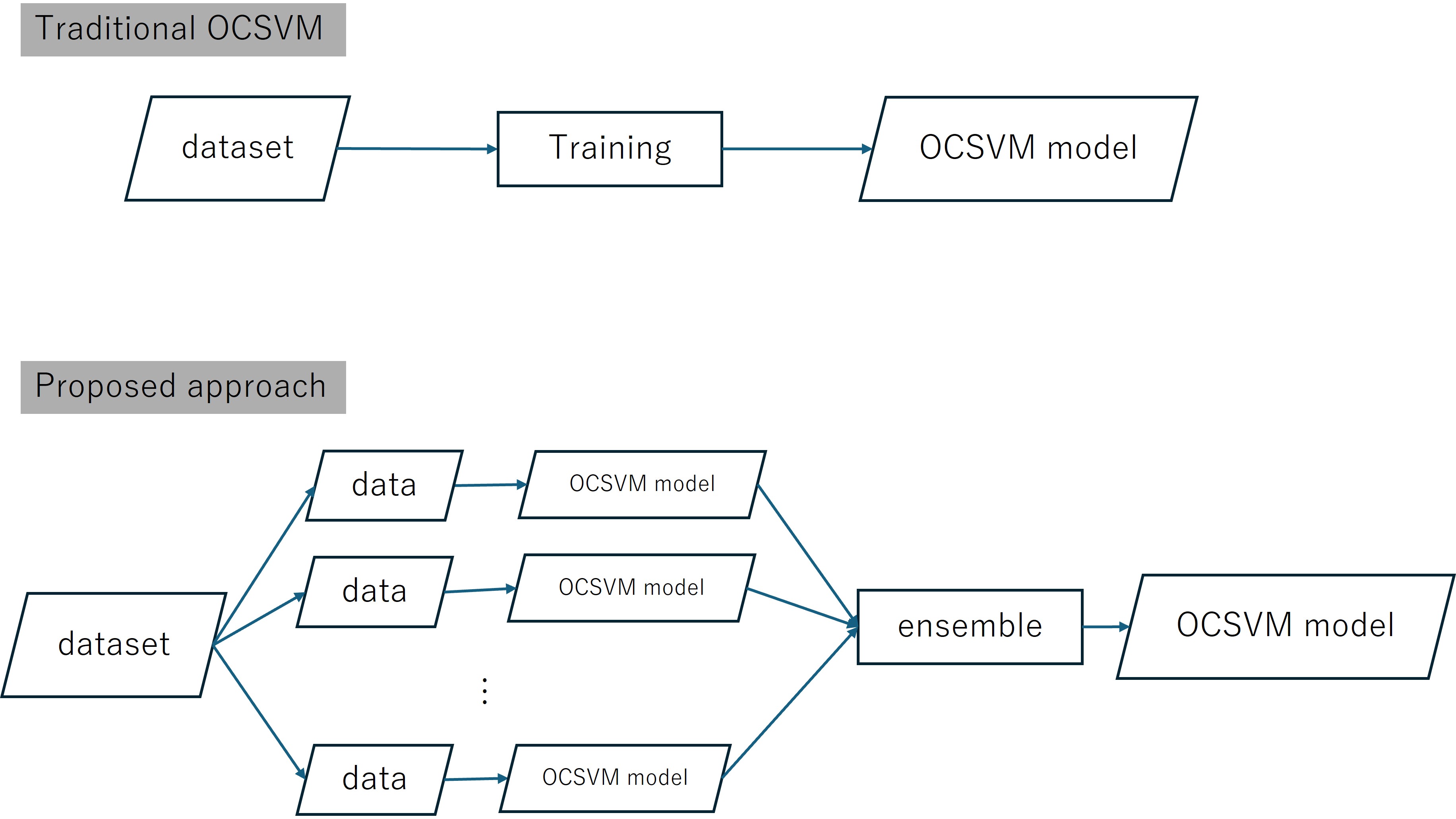}
\caption{ODSVM ensemble}\label{fig1}
\end{figure}

This paper proposes an acceleration strategy for OCSVM. Figure \ref{fig1} shows the processing image: the traditional algorithms train an OCSVM model from the entire dataset. However, such a process requires optimization for a large-scale dataset. On the other hand, this paper decomposes the dataset into samples and trains OCSVM models for each sample, namely ODSVM models. Subsequently, ensemble learning is applied to combine the ODSVM models into an OCSVM model for the dataset. 

\subsection{One-data support vector machine}
\label{sec3.1}
ODSVM is an OCSVM that learns from a single sample. When $n$ = 1, the training of OCSVM (equation (\ref{eq8})) is simplified into equation (\ref{eq11}):
\begin{equation}\label{eq11}
⁡\frac{1}{2} K(x_{tr},x_{tr}),
\end{equation}
where $x_{tr}$ refers to a training sample. In addition, equation (\ref{eq12}) shows the score of ODSVM, which is the simplified equation (\ref{eq3}):
⁡\begin{equation}\label{eq12}
⁡{score}_{ODSVM}(x)=K(x_{tr},x)-\rho.
\end{equation}

The score of OCSVM could be approximated by ensemble learning of ODSVM models, as shown in equation (\ref{eq13}):
\begin{equation}\label{eq13}
⁡score_{OCSVM}(x) \approx ensemble({score}_{ODSVM}(x)).
\end{equation}
The detailed ensemble process is described in subsection \ref{sec3.3} and equations (\ref{eq14})-(\ref{eq15}). The next subsections describe training and testing stages with pseudocode (Python).

\subsection{Training stages}\label{sec3.2}
The training stage aims to create ODSVM models from the training set. This subsection proposes two algorithms. Algorithm \ref{alg1} shows the pseudocode of ODSVM-all, which is a straightforward approach to train ODSVM models from all samples. 
\begin{algorithm}[H]
\caption{ODSVM-all (Training ODSVM models for all data)}\label{alg1}
\textbf{Input}: training dataset

\textbf{Output}: models

models= []

\textbf{For} data in dataset:

\quad Initialize OCSVM // ODSVM 

\quad OCSVM.fit(data) // data.reshape(1, -1)

\quad models.append(OCSVM)

\end{algorithm}

However, this algorithm is time-consuming in the testing stage because the number of OCSVM models equals the number of training samples. Therefore, the number of models should be reduced. For this purpose, revisiting the principle of SVM is a reasonable direction. In the original SVM, support vectors refer to samples around the borderline. Similarly, support vectors in OCSVM are the most abnormal training samples because they are the closest to the boundary. Therefore, only abnormal training samples are important to approximate OCSVM models.

Accordingly, algorithm \ref{alg2} proposes a training process with a model reduction strategy. The proposed strategy trains an ODSVM model on an average of training samples and computes normal scores for the entire dataset. Subsequently, N abnormal samples (low normal scores) are selected as support vectors to train ODSVM models, where N is a hyperparameter referring to the number of support vectors.

\begin{algorithm}[H]
\caption{ODSVM-N (Training N ODSVM models)}\label{alg2}
\textbf{Input}: dataset, and number of models N

\textbf{Output}: N models

//Data Reduction (Select N support vectors)

Initialize OCSVM //ODSVM

OCSVM.fit(dataset.mean(axis=0)) 

score=OCSVM.score\_samples(dataset) //Normal score

rank=score.argsort().argsort() //ascending order (small number first)

//Support vectors of OCSVM are the most abnormal training samples.

Support=dataset[rank $<$ N] 

//Training only support vectors.

models = []

\textbf{For} data in Support:

\quad    Initialize OCSVM //ODSVM

\quad	OCSVM.fit(data)

\quad	models.append(OCSVM)

\end{algorithm}

\subsection{Testing stages}
\label{sec3.3}
The testing stage combines the score of ODSVM models. This paper applies two simple ensemble strategies, 1) summation, and 2) using the maximum score. 

The first strategy is summation of scores:
\begin{equation}\label{eq14}
⁡score_{OCSVM}(x)=\sum_{i=1}^N ⁡score_{ODSVMi}(x),
\end{equation}
where N is number of models. Note that normalization is not necessary because the kernel function performs as a normalization. Algorithm \ref{alg3} shows pseudocode of summation.  

\begin{algorithm}[H]
\caption{Summation of normal scores}\label{alg3}

\textbf{Input}: Testing dataset X\_test, models

\textbf{Output}: ensemble score Ens\_score

Ens\_score=np.zeros(len(X\_test))

\textbf{For} model in models:

\quad Ens\_score+=model.score\_samples(X\_test)
\end{algorithm}
Another strategy is to select single model output, which is the largest score among all ODSVM models:
\begin{equation}\label{eq15}
⁡score_{OCSVM}(x) = \max_{i}(⁡score_{ODSVMi}(x)).
\end{equation}
The score is provided from the ODSVM model trained on the nearest training sample (or support vector) to the testing sample $x$. 
Algorithm \ref{alg4} presents the pseudocode for computing the maximum scores.

\begin{algorithm}[H]
\caption{Output the maximum scores}\label{alg4}

\textbf{Input}: Testing dataset X\_test, models

\textbf{Output}: ensemble scores

scores=[]

\textbf{For} model in models:

\quad scores.append(model.score\_samples(X\_test))

ens\_score=np.array(scores).max(axis=0)

\end{algorithm}

In summary, this section proposed four possible pairs of training and testing algorithms, as shown in table \ref{tableN1}. Two pairs (algorithms \ref{alg2}+\ref{alg3} and \ref{alg2}+\ref{alg4}) are new proposals. On the other hand, algorithm \ref{alg1} + \ref{alg3} is equivalent to kernel density estimation. Moreover, algorithm \ref{alg4} should not be combined with algorithm \ref{alg1} because the space complexity is high.

\begin{table}[H]
\caption{Summary of the proposals}\label{tableN1}
\centering
\scalebox{1}{
\begin{tabular}{l | l | l } \hline
Training & Testing  & Other information \\ \hline
Algorithm 1 & Algorithm 3 & Theoretically equivalent to kernel density estimation \cite{KDE}. \\
Algorithm 1 & Algorithm 4 & There is a scalability problem for space complexity. \\
Algorithm 2 & Algorithm 3 & New proposal \\
Algorithm 2 & Algorithm 4 & New proposal \\ \hline
\end{tabular}
}
\end{table}

\section{Experiment}
\label{sec4}
The experiment compares the original OCSVM and the proposed decomposition strategies. The evaluation criteria are Area Under the ROC Curve (AUROC), processing time, and file size. Subsection \ref{sec4.1} describes the datasets, while subsections \ref{sec4.2} and \ref{sec4.3} report AUROC and processing time, respectively. Subsection \ref{sec4.4} analyzes the file sizes of models. In addition, subsection \ref{sec4.5} applies ODSVM to the time series dataset. 

\subsection{Dataset}
\label{sec4.1}
The experiment uses an imbalanced learn dataset \cite{[14]} that contains 27 binary classification datasets. Table \ref{table1} shows the number of minority and majority samples and their dimensions.

\begin{table}[H]
\caption{Characteristics of datasets}\label{table1}
\centering
\scalebox{1}{
\begin{tabular}{l l l l l l l l} \hline
Dataset & Minority & Majority & Dimension \\ \hline
Ecoli & 35 & 301 & 7 \\ 
Optical\_digits & 554 & 5066 & 64 \\ 
Satimage & 626 & 5809 & 36 \\ 
Pen\_digits & 1055 & 9937 & 16 \\ 
Abalone & 391 & 3786 & 10 \\ 
Sick\_euthyroid & 293 & 2870 & 42 \\ 
Spectrometer & 45 & 486 & 93 \\ 
Car\_eval\_34 & 134 & 1594 & 21 \\ 
Isolet & 600 & 7197 & 617 \\ 
Us\_crime & 150 & 1844 & 100 \\ 
Yeast\_ml8 & 178 & 2239 & 103 \\ 
Scene & 177 & 2230 & 294 \\ 
Libras\_move & 24 & 336 & 90 \\ 
Thyroid\_sick & 231 & 3541 & 52 \\ 
Coil\_200 & 586 & 9236 & 85 \\ 
Arrhythmia & 25 & 427 & 278 \\ 
Solar\_flare\_m0 & 68 & 1321 & 32 \\ 
Oil & 41 & 896 & 49 \\ 
Car\_eval\_4 & 65 & 1663 & 21 \\ 
Wine\_quality & 183 & 4715 & 11 \\ 
Letter\_img & 734 & 19266 & 16 \\ 
Yeast\_me2 & 51 & 1433 & 8 \\ 
Webpage & 981 & 33799 & 300 \\ 
Ozone\_level & 73 & 2463 & 72 \\ 
Mammography & 260 & 10923 & 6 \\ 
Protein\_homo & 1296 & 144455 & 74 \\
Abalone\_19 & 32 & 4145 & 10 \\ \hline
\end{tabular}
}
\end{table}

\subsection{Area under ROC curve}
\label{sec4.2}
This section compares the proposed decomposition methods and the original OCSVM. Both ODSVM and baseline OCSVM use a sklearn package \cite{[6]}, where the hyperparameter is default. ODSVM-all refers to an ensemble of all ODSVM models computed by Algorithm \ref{alg1}. ODSVM-200 refers to an ensemble of 200 ODSVM models, while ODSVM-1000 is an ensemble of 1000 ODSVM models. These methods are computed by Algorithm \ref{alg2}. Note that some settings have less than 200 or 1000 training data. In such a case, the result is equivalent to ODSVM-all.

Table \ref{table2} compares AUC scores for OCSVM, ODSVM-all, ODSVM-200, and ODSVM-1000. ODSVM shows similar AUC to the original OCSVM. This result supports that decomposing OCSVM into an ensemble of ODSVMs is a reasonable approximation process.

\begin{table}[H]
\caption{Comparison between OCSVM, ODSVM-all, and ODSVM-N (N=200 and 1000)}\label{table2}
\centering
\scalebox{0.6}{
\begin{tabular}{l l l l l l l | l l l l l l} \hline
 & \multicolumn{6}{l|}{One = minority class} &  \multicolumn{6}{l}{One = majority class} \\ \hline
Dataset & OCSVM & ODSVM-all & \multicolumn{2}{l}{ODSVM-200} & \multicolumn{2}{l|}{ODSVM-1000} & OCSVM & ODSVM-all & \multicolumn{2}{l}{ODSVM-200} & \multicolumn{2}{l}{ODSVM-1000} \\
  &  & SUM & SUM & MAX & SUM & MAX &  & SUM & SUM & MAX & SUM & MAX \\ \hline
Ecoli & 91.5 & 91.7 & 91.7 & 91.6 & 91.7 & 91.6 & 77.0 & 82.5 & 82.5 & 41.7 & 82.5 & 41.7 \\ 
optical\_digits & 97.6 & 97.5 & 97.6 & 97.6 & 97.5 & 98.4 & 21.6 & 25.1 & 27.1 & 85.2 & 22.6 & 90.9 \\ 
Satimage & 91.5 & 91.6 & 89.7 & 79.8 & 91.6 & 87.7 & 23.1 & 33.1 & 70.1 & 75.2 & 65.4 & 74.0 \\ 
pen\_digits & 99.0 & 97.2 & 40.2 & 41.0 & 97.2 & 99.8 & 78.6 & 77.3 & 45.6 & 47.9 & 45.5 & 52.8 \\ 
Abalone & 81.8 & 80.7 & 80.6 & 81.2 & 80.7 & 82.8 & 50.0 & 65.6 & 35.3 & 35.8 & 30.7 & 27.1 \\ 
sick\_euthyroid & 69.5 & 68.1 & 68.1 & 83.8 & 68.1 & 83.8 & 43.6 & 39.7 & 65.5 & 64.6 & 43.9 & 51.1 \\
Spectrometer & 52.7 & 52.1 & 52.1 & 63.4 & 52.1 & 63.4 & 82.4 & 79.8 & 82.0 & 73.4 & 79.8 & 78.6 \\ 
car\_eval\_34 & 99.8 & 98.9 & 98.9 & 91.1 & 98.9 & 91.1 & 97.7 & 96.7 & 82.7 & 61.8 & 96.7 & 63.7 \\ 
Isolet & 93.0 & 92.5 & 93.0 & 94.6 & 92.5 & 95.4 & 22.3 & 20.1 & 59.8 & 75.4 & 41.7 & 68.6 \\ 
us\_crime & 73.4 & 69.6 & 69.6 & 73.0 & 69.6 & 73.0 & 78.9 & 80.2 & 72.1 & 51.6 & 80.1 & 76.1 \\ 
yeast\_ml8 & 56.3 & 58.4 & 58.4 & 51.4 & 58.4 & 51.4 & 49.4 & 48.4 & 54.8 & 57.0 & 51.9 & 57.5 \\ 
Scene & 62.6 & 61.5 & 61.5 & 62.6 & 61.5 & 62.6 & 49.1 & 48.6 & 57.9 & 44.8 & 48.7 & 46.1 \\ 
libras\_move & 90.5 & 89.7 & 89.7 & 96.2 & 89.7 & 96.2 & 83.1 & 73.8 & 73.7 & 92.0 & 73.8 & 92.1 \\ 
thyroid\_sick & 71.7 & 73.0 & 73.0 & 78.3 & 73.0 & 78.3 & 52.8 & 47.2 & 73.1 & 69.7 & 55.8 & 56.6 \\ 
coil\_2000 & 54.5 & 54.6 & 54.6 & 57.2 & 54.6 & 58.5 & 54.2 & 55.4 & 43.0 & 44.6 & 47.7 & 46.8 \\ 
Arrhythmia & 63.1 & 61.0 & 61.0 & 67.6 & 61.0 & 67.6 & 44.2 & 43.4 & 43.6 & 40.9 & 43.4 & 44.3 \\ 
solar\_flare\_m0 & 56.7 & 52.7 & 52.7 & 52.4 & 52.7 & 52.4 & 75.7 & 78.1 & 58.2 & 42.1 & 78.1 & 67.5 \\ 
Oil & 63.6 & 57.6 & 57.6 & 68.5 & 57.6 & 68.5 & 62.7 & 64.3 & 56.5 & 65.3 & 64.3 & 85.8 \\ 
car\_eval\_4 & 99.9 & 99.6 & 99.6 & 94.2 & 99.6 & 94.2 & 99.1 & 97.1 & 74.1 & 56.1 & 97.1 & 55.9 \\ 
wine\_quality & 49.3 & 48.4 & 48.4 & 59.4 & 48.4 & 59.4 & 65.4 & 64.2 & 65.6 & 54.6 & 68.2 & 61.2 \\ 
letter\_img & 89.7 & 88.4 & 88.7 & 97.3 & 88.4 & 99.3 & 60.2 & 54.3 & 73.2 & 67.1 & 78.0 & 65.1 \\ 
yeast\_me2 & 79.4 & 78.5 & 78.5 & 84.4 & 78.5 & 84.4 & 74.9 & 75.9 & 69.6 & 25.2 & 75.9 & 62.0 \\ 
Webpage & 62.2 & 61.7 & 62.2 & 82.6 & 61.7 & 87.7 & 49.2 & 47.4 & 82.4 & 77.0 & 76.4 & 89.1 \\ 
ozone\_level & 78.5 & 79.9 & 79.9 & 81.5 & 79.9 & 81.5 & 52.8 & 42.9 & 79.5 & 24.9 & 40.9 & 24.3 \\ 
Mammography & 79.7 & 80.0 & 80.0 & 47.9 & 80.0 & 47.9 & 85.3 & 59.3 & 63.2 & 23.2 & 37.4 & 28.2 \\ 
protein\_homo & 80.8 & 78.0 & 88.0 & 38.1 & 78.0 & 70.9 & 86.1 & 84.8 & 68.1 & 43.6 & 64.3 & 68.8 \\ 
abalone\_19 & 69.7 & 68.0 & 68.0 & 63.2 & 68.0 & 63.2 & 38.4 & 40.4 & 56.7 & 52.7 & 60.1 & 64.4 \\ 
Average & 76.2 & 75.2 & 73.4 & 73.3 & 75.2 & 77.4 & 61.4 & 60.2 & 63.5 & 55.3 & 61.1 & 60.8 \\ \hline
\end{tabular}
}
\end{table}

Tables \ref{table3} and \ref{table4} compare the ODSVM with other acceleration strategies as follows:

\textbf{OCSVM}\cite{[1]} is a sklearn baseline \cite{[6]} using default hyperparameter. 

\textbf{SGD-OCSVM} is an OCSVM variant that uses stochastic gradient descent (SGD). SGD minimizes the loss using only a few random samples. This strategy is significantly faster than optimizing with the entire dataset. The kernel is linear.

\textbf{Nystrom kernel \cite{NystromSVM}} is the technique to minimize kernel matrix. The idea is to create a matrix using only a few random samples. This method is combined with SGD-OCSVM.

\textbf{N-OCSVM \cite{subsetSVM}} is an ensemble of OCSVM models trained on N samples. The experiment used N=2, 10, 100, and 1000. Note that N=1 is equivalent to ODSVM-all. The subsets are randomly created by indexed order.

\textbf{Krawczyk et al. \cite{[2]}} combined K-means and ensemble of OCSVM models. The idea was to apply K-means to the dataset and train OCSVM models from each cluster. Finally, ensemble learning is applied to combine the OCSVM models. This experiment uses K=10.

Table \ref{table3} compares AUC scores, where one class is minority. ODSVM-1000 MAX is the best in the average AUC scores. 

\begin{table}[H]
\caption{Comparison with other acceleration strategy (One class is minority)}\label{table3}
\centering
\scalebox{0.6}{
\begin{tabular}{l | l | l l | l l  l  l |l | l | l  l | l l } \hline
 & Original & & & \multicolumn{4}{l|}{N-OCSVM} & Krawczyk & ODSVM & \multicolumn{2}{l|}{ODSVM-200} & \multicolumn{2}{l}{ODSVM-1000}\\
One=minority & OCSVM & SGD & Nystrom+SGD & 2 & 10 & 100 & 1000 & k=10 & all SUM & SUM & MAX & SUM & MAX \\ \hline
Ecoli & 91.5 & 91.8 & 90.6 & 91.7 & 91.1 & 91.5 & 91.5 & 91.6 & 91.7 & 91.7 & 91.6 & 91.7 & 91.6 \\
optical\_digits & 97.6 & 84.9 & 97.3 & 97.5 & 97.6 & 97.6 & 97.6 & 97.5 & 97.5 & 97.6 & 97.6 & 97.5 & 98.4 \\
Satimage & 91.5 & 58.7 & 91.7 & 91.7 & 91.6 & 91.5 & 91.5 & 91.7 & 91.6 & 89.7 & 79.8 & 91.6 & 87.7 \\
pen\_digits & 99 & 62.1 & 71.1 & 97.2 & 99.2 & 99 & 99 & 97 & 97.2 & 40.2 & 41 & 97.2 & 99.8 \\
Abalone & 81.8 & 34.5 & 81.9 & 80.7 & 80.5 & 81.4 & 81.8 & 80.6 & 80.7 & 80.6 & 81.2 & 80.7 & 82.8 \\
sick\_euthyroid & 69.5 & 65 & 69 & 68.1 & 68.8 & 69.3 & 69.5 & 67.7 & 68.1 & 68.1 & 83.8 & 68.1 & 83.8 \\
Spectrometer & 52.7 & 56.9 & 27.1 & 50.6 & 55 & 52.7 & 52.7 & 51.4 & 52.1 & 52.1 & 63.4 & 52.1 & 63.4 \\
car\_eval\_34 & 99.8 & 99.6 & 99.7 & 98.9 & 99.5 & 99.8 & 99.8 & 99.2 & 98.9 & 98.9 & 91.1 & 98.9 & 91.1 \\
Isolet & 93 & 67.4 & 92.5 & 92.5 & 93 & 93.1 & 93 & 92.6 & 92.5 & 93 & 94.6 & 92.5 & 95.4 \\
us\_crime & 73.4 & 72.4 & 73.3 & 69.4 & 72.1 & 73.4 & 73.4 & 69.6 & 69.6 & 69.6 & 73 & 69.6 & 73.0 \\
yeast\_ml8 & 56.3 & 44.6 & 57.9 & 58.3 & 58.1 & 56.6 & 56.3 & 57.7 & 58.4 & 58.4 & 51.4 & 58.4 & 51.4 \\
Scene & 62.6 & 62 & 63 & 61.7 & 62.1 & 62.5 & 62.6 & 62.6 & 61.5 & 61.5 & 62.6 & 61.5 & 62.6 \\
libras\_move & 90.5 & 54 & 73.9 & 88.7 & 87.4 & 90.5 & 90.5 & 89.5 & 89.7 & 89.7 & 96.2 & 89.7 & 96.2 \\
thyroid\_sick & 71.7 & 72.7 & 72.5 & 73 & 73 & 71.9 & 71.7 & 72.5 & 73 & 73 & 78.3 & 73.0 & 78.3 \\
coil\_2000 & 54.5 & 66.7 & 54.5 & 54.7 & 54.4 & 54.4 & 54.5 & 54.1 & 54.6 & 54.6 & 57.2 & 54.6 & 58.5 \\
Arrhythmia & 63.1 & 61.3 & 61.5 & 60.5 & 62.5 & 63.1 & 63.1 & 60.8 & 61 & 61 & 67.6 & 61.0 & 67.6 \\
solar\_flare\_m0 & 56.7 & 61.3 & 57.2 & 53 & 55.5 & 56.7 & 56.7 & 53.4 & 52.7 & 52.7 & 52.4 & 52.7 & 52.4 \\
Oil & 63.6 & 56.5 & 62.5 & 57.2 & 61.4 & 63.6 & 63.6 & 57.9 & 57.6 & 57.6 & 68.5 & 57.6 & 68.5 \\
car\_eval\_4 & 99.9 & 99.7 & 99.8 & 99.6 & 99.6 & 99.9 & 99.9 & 99.6 & 99.6 & 99.6 & 94.2 & 99.6 & 94.2 \\
wine\_quality & 49.3 & 50.4 & 48.7 & 48.4 & 48.6 & 49.1 & 49.3 & 48.8 & 48.4 & 48.4 & 59.4 & 48.4 & 59.4 \\
letter\_img & 89.7 & 72.2 & 85 & 88.3 & 89.4 & 89.7 & 89.7 & 88.4 & 88.4 & 88.7 & 97.3 & 88.4 & 99.3 \\
yeast\_me2 & 79.4 & 78.5 & 77.5 & 79.8 & 79.4 & 79.4 & 79.4 & 79.1 & 78.5 & 78.5 & 84.4 & 78.5 & 84.4 \\
Webpage & 62.2 & 56.6 & 60.9 & 61 & 61.7 & 62.2 & 62.2 & 61.5 & 61.7 & 62.2 & 82.6 & 61.7 & 87.7 \\
ozone\_level & 78.5 & 70.6 & 76.9 & 80 & 79.1 & 78.5 & 78.5 & 79.5 & 79.9 & 79.9 & 81.5 & 79.9 & 81.5 \\
Mammography & 79.7 & 88.5 & 72.1 & 79.7 & 79.4 & 79.6 & 79.7 & 80.4 & 80 & 80 & 47.9 & 80.0 & 47.9 \\
protein\_homo & 80.8 & 90.6 & 83 & 77.4 & 79.9 & 80.5 & 80.8 & 78.1 & 78 & 88 & 38.1 & 78.0 & 70.9 \\
abalone\_19 & 69.7 & 68 & 69.3 & 68.1 & 71 & 69.7 & 69.7 & 67.9 & 68 & 68 & 63.2 & 68.0 & 63.2 \\
Average & 76.2 & 68.4 & 73 & 75.1  & 76.0  & 76.2  & 76.2  & 75.2 & 75.2 & 73.4 & 73.3 & 75.2 & 77.4 \\ \hline
\end{tabular}
}
\end{table}

Table \ref{table4} compares AUC scores, where one class is majority. ODSVM-200 SUM is the best in average AUC score.
\begin{table}[H]
\caption{Comparison with other acceleration strategy (Majority) }\label{table4}
\centering
\scalebox{0.6}{
\begin{tabular}{l l | l l | l l l  l | l | l | l l| l  l } \hline
 & Original & & & \multicolumn{4}{l|}{N-OCSVM} & Krawczyk & ODSVM &  \multicolumn{2}{l|}{ODSVM-200} &  \multicolumn{2}{l}{ODSVM-1000} \\
One=majority & OCSVM & SGD & Nystrom+SGD & 2 & 10 & 100 & 1000 & k=10 & all SUM & SUM & MAX & SUM & MAX \\ \hline
Ecoli & 77 & 10.4 & 76 & 82.8 & 76 & 77.5 & 77 & 82.2 & 82.5 & 82.5 & 41.7 & 82.5 & 41.7 \\
optical\_digits & 21.6 & 27.2 & 16.6 & 24.9 & 16.6 & 21.8 & 21.7 & 23.8 & 25.1 & 27.1 & 85.2 & 22.6 & 90.9 \\
Satimage & 23.1 & 43.8 & 19.9 & 16.9 & 19.9 & 22.5 & 23 & 25.3 & 33.1 & 70.1 & 75.2 & 65.4 & 74.0 \\
pen\_digits & 78.6 & 61.5 & 77.4 & 77.1 & 77.4 & 78.7 & 78.7 & 76.1 & 77.3 & 45.6 & 47.9 & 45.5 & 52.8 \\
Abalone & 50 & 77.4 & 61.2 & 66.1 & 61.2 & 50.2 & 50.1 & 65.8 & 65.6 & 35.3 & 35.8 & 30.7 & 27.1 \\
sick\_euthyroid & 43.6 & 45.9 & 44.1 & 39.7 & 44.1 & 43.3 & 43.6 & 39.9 & 39.7 & 65.5 & 64.6 & 43.9 & 51.1 \\
Spectrometer & 82.4 & 58.3 & 81.1 & 80 & 81.1 & 82.2 & 82.4 & 79.5 & 79.8 & 82.0 & 73.4 & 79.8 & 78.6 \\
car\_eval\_34 & 97.7 & 60.2 & 88.3 & 96.7 & 88.3 & 96.9 & 97.7 & 86.2 & 96.7 & 82.7 & 61.8 & 96.7 & 63.7 \\
Isolet & 22.3 & 45.1 & 21.2 & 20 & 21.2 & 22.2 & 22.3 & 18.7 & 20.1 & 59.8 & 75.4 & 41.7 & 68.6 \\
us\_crime & 78.9 & 59.5 & 78.1 & 80.1 & 78.1 & 78.9 & 78.9 & 79.9 & 80.2 & 72.1 & 51.6 & 80.1 & 76.1 \\
yeast\_ml8 & 49.4 & 55.5 & 47.7 & 48.4 & 47.7 & 50 & 49.4 & 47.9 & 48.4 & 54.8 & 57.0 & 51.9 & 57.5 \\
Scene & 49.1 & 43.7 & 49.0 & 48.4 & 49.0 & 49.0 & 49.1 & 48.2 & 48.6 & 57.9 & 44.8 & 48.7 & 46.1 \\
libras\_move & 83.1 & 57.6 & 59.1 & 71.7 & 59.1 & 83.2 & 83.1 & 71.5 & 73.8 & 73.7 & 92.0 & 73.8 & 92.1 \\
thyroid\_sick & 52.8 & 54.6 & 53 & 47.2 & 53 & 52.5 & 52.8 & 46.8 & 47.2 & 73.1 & 69.7 & 55.8 & 56.6 \\
coil\_2000 & 54.2 & 38.8 & 54.3 & 55.4 & 54.3 & 54.3 & 54.2 & 55.1 & 55.4 & 43 & 44.6 & 47.7 & 46.8 \\
Arrhythmia & 44.2 & 43.2 & 44.4 & 43.3 & 44.4 & 44.4 & 44.2 & 44.0 & 43.4 & 43.6 & 40.9 & 43.4 & 44.3 \\
solar\_flare\_m0 & 75.7 & 76 & 76.8 & 78.1 & 76.8 & 76.8 & 75.7 & 77.2 & 78.1 & 58.2 & 42.1 & 78.1 & 67.5 \\
Oil & 62.7 & 55.6 & 52.6 & 64.3 & 52.6 & 62.9 & 62.7 & 63.2 & 64.3 & 56.5 & 65.3 & 64.3 & 85.8 \\
car\_eval\_4 & 99.1 & 52.5 & 79 & 97.1 & 79 & 95.9 & 99.1 & 93.2 & 97.1 & 74.1 & 56.1 & 97.1 & 55.9 \\
wine\_quality & 65.4 & 53.4 & 62.8 & 64 & 62.8 & 65.4 & 65.4 & 63.6 & 64.2 & 65.6 & 54.6 & 68.2 & 61.2 \\
letter\_img & 60.2 & 41.9 & 58.4 & 53.9 & 58.4 & 59.9 & 60.2 & 55.7 & 54.3 & 73.2 & 67.1 & 78.0 & 65.1 \\
yeast\_me2 & 74.9 & 28.5 & 75.3 & 75.6 & 75.3 & 74.8 & 74.9 & 76.3 & 75.9 & 69.6 & 25.2 & 75.9 & 62.0 \\
Webpage & 49.2 & 72.7 & 49.5 & 46.2 & 49.5 & 49.1 & 49.2 & 48.1 & 47.4 & 82.4 & 77.0 & 76.4 & 89.1 \\
ozone\_level & 52.8 & 38.3 & 55.5 & 43.8 & 55.5 & 52.5 & 52.7 & 42.9 & 42.9 & 79.5 & 24.9 & 40.9 & 24.3 \\
Mammography & 85.3 & 23.7 & 84.8 & 84.9 & 84.8 & 85.4 & 85.3 & 76.4 & 59.3 & 63.2 & 23.2 & 37.4 & 28.2 \\
protein\_homo & 86.1 & 17.6 & 86 & 84.8 & 86 & 86.1 & 86.1 & 85 & 84.8 & 68.1 & 43.6 & 64.3 & 68.8 \\
abalone\_19 & 38.4 & 36.3 & 42.3 & 40.2 & 42.3 & 38.7 & 38.7 & 40.3 & 40.4 & 56.7 & 52.7 & 60.1 & 64.4 \\
Average & 61.4 & 47.3 & 59.0 & 60.4  & 59.1  & 61.3  & 61.4  & 59.7 & 60.2 & 63.5 & 55.3 & 61.1 & 60.8 \\ \hline
\end{tabular}
}
\end{table}

\subsection{Processing speed}
\label{sec4.3}
This section compares processing time of the traditional OCSVM and the proposed ODSVM. Both OCSVM and ODSVM used sklearn package \cite{[6]}. Tables \ref{table5} and \ref{table6} report the processing time for protein\_homo and webpage datasets, respectively. Each table compares the OCSVM, ODSVM-all, ODSVM-200, and ODSVM-1000.  
Decomposing OCSVM can reduce computational cost for large datasets. However, ODSVM-all takes a longer time in the testing stage, because the number of ODSVM models equals the number of training samples. Accordingly, the number of models should be reduced to reasonable sizes, such as 200 or 1000. ODSVM-200 and 1000 successfully reduced the processing time of OCSVM, while maintaining AUC scores. Therefore, an ensemble of ODSVM can be a strong alternative to OCSVM variants. 

Table \ref{table5} reports the processing time of the protein homo dataset, where the training set contains 86672 samples, while the testing set contains 58301 samples. Summation (Algorithm \ref{alg3}) and max(Algorithm \ref{alg4}) show similar processing time, while Algorithm \ref{alg4} requires more memory space.

\begin{table}[H]
\caption{Processing time (One class = protein homo majority)}\label{table5}
\centering
\scalebox{0.9}{
\begin{tabular}{l l l l l l l l} \hline
Method & Training & Testing & Total \\ 
& (86672 samples) & (58301 samples) \\ \hline
OCSVM & 7 min 57 s & 2 min 31 s & 10 min 28 s \\
K-means (K=10) + OCSVM ensemble & 23 s & 1 min 48 s & 2 min 38 s\\
SGD-OCSVM & 100 ms & 72.5 ms & 172.5 ms \\
Nystromem + SGD-OCSVM & 253 ms & 812 ms & 1.07 s \\
Ensemble of OCSVMs for 2 samples & 13.9 s & 6 min 30 s &  6 min 44 s \\
Ensemble of OCSVMs for 10 samples & 2.93 s & 3 min 5 s & 3 min 8 s \\
Ensemble of OCSVMs for 100 samples & 531 ms & 1 min 51 s & 1 min 52 s \\
Ensemble of OCSVMs for 1000 samples & 1.95 s & 1 min 46 s &  1 min 48 s \\ \hline
ODSVM-all & 29 s & 8 min 55 s & 10 min 24 s \\
ODSVM-200-sum & 271 ms & 1.47 s & 1.7 s \\
ODSVM-200-max & 271 ms & 1.42 s & 1.7 s \\
ODSVM-1000-sum & 525 ms & 6.84 s & 7.4 s \\
ODSVM-1000-max & 525 ms & 7.03 s & 7.6 s \\ \hline
\end{tabular}
}
\end{table}

Table \ref{table6} reports the processing time of the webpage dataset, where the training set contains 20,279 samples, while the testing set contains 13,912 samples. The reported result is an ensemble using the summation. Training and testing times are related to the number of ODSVM models. ODSVM-N successfully accelerated the processing speed from original OCSVM models. SGD-OCSVM and Nystrom kernel is faster than OCSVM.

SGD-OCSVM use linear kernel. Therefore, its model is generally nonsense. Nystrom computes RBF kernel with smaller samples. However, the algorithm still solves small optimization problem. Therefore, the result might have error. On the other hand, ODSVM does not solve optimization problem. The model is computed by combination of obvious solutions.

\begin{table}[H]
\caption{Processing time (One class = webpage majority)}\label{table6}
\centering
\scalebox{0.9}{
\begin{tabular}{l l l l l l l l} \hline
Method & Training & Testing  & Total \\
& (20279 samples) & (13912 samples) \\ \hline
OCSVM & 58.4 s & 30 s & 1 min 26 s \\ 
K-means (K=10) + OCSVM ensemble & 3.08 s & 20.4 s & 23.5 s\\
SGD-OCSVM & 66.6 ms & 6.12 ms & 72.7 ms \\
Nystromem + SGD-OCSVM & 79.6 ms & 34.4 ms & 114 ms \\
Ensemble of OCSVMs for 2 samples & 3.16 s & 1 min 19 s &  1 min 22 s \\
Ensemble of OCSVMs for 10 samples & 697 ms & 43.2 s & 43.9 s \\
Ensemble of OCSVMs for 100 samples & 176 ms & 23.1 s & 23.3  s \\
Ensemble of OCSVMs for 1000 samples & 960 ms & 20 s & 20.9 s \\ \hline
ODSVM-all & 7.1 s & 2 min 3 s & 2 min 10 s \\ 
ODSVM-200 & 228 ms & 1.2 s & 1.4 s \\ 
ODSVM-1000 & 500 ms & 5.99 s & 6.5 s \\ \hline
\end{tabular}
}
\end{table}

Training on multiple samples is complex because the model has to be balanced across all samples. Optimization has a risk of failure. On the other hand, the model from a single sample does not need such a balance.

\subsection{The memory space}
\label{sec4.4}
This subsection analyzes space complexity. For this purpose, Table \ref{table7} compares the file sizes of the training dataset, traditional OCSVM model, and the proposed ODSVM models. The experiment used the majority class of the protein\_homo dataset because it is the largest dataset in imbalanced-learn. All objects are saved as pickle files for measuring. 

The original training dataset has 48.9 MB, while the traditional OCSVM model is 25 MB. On the other hand, ODSVM-all created 86,672 models, using 92.8 MB in total. Such a size is approximately twice of the training dataset. In contrast, ODSVM-200 is 219.9 KB, and ODSVM-1000 is 1.1 MB. Accordingly, ODSVM-N can reduce the memory sizes compared to the traditional OCSVM.

\begin{table}[H]
\caption{File size (One class = protein\_homo majority)}\label{table7}
\centering
\scalebox{0.9}{
\begin{tabular}{l l } \hline
Object & File size \\ \hline
Training set (86672 samples) & 48.9 MB  \\
Traditional OCSVM model & 25 MB  \\
ODSVM-all (86672 models) & 92.8 MB \\
ODSVM-200 (200 models) & 219.9 KB  \\ 
ODSVM-1000 (1000 models) & 1.1 MB \\ \hline
\end{tabular}
}

\end{table}

\subsection{Applying ODSVM to time-series dataset}
\label{sec4.5}
\begin{figure}[H]
\centering
\includegraphics[width=1\linewidth]{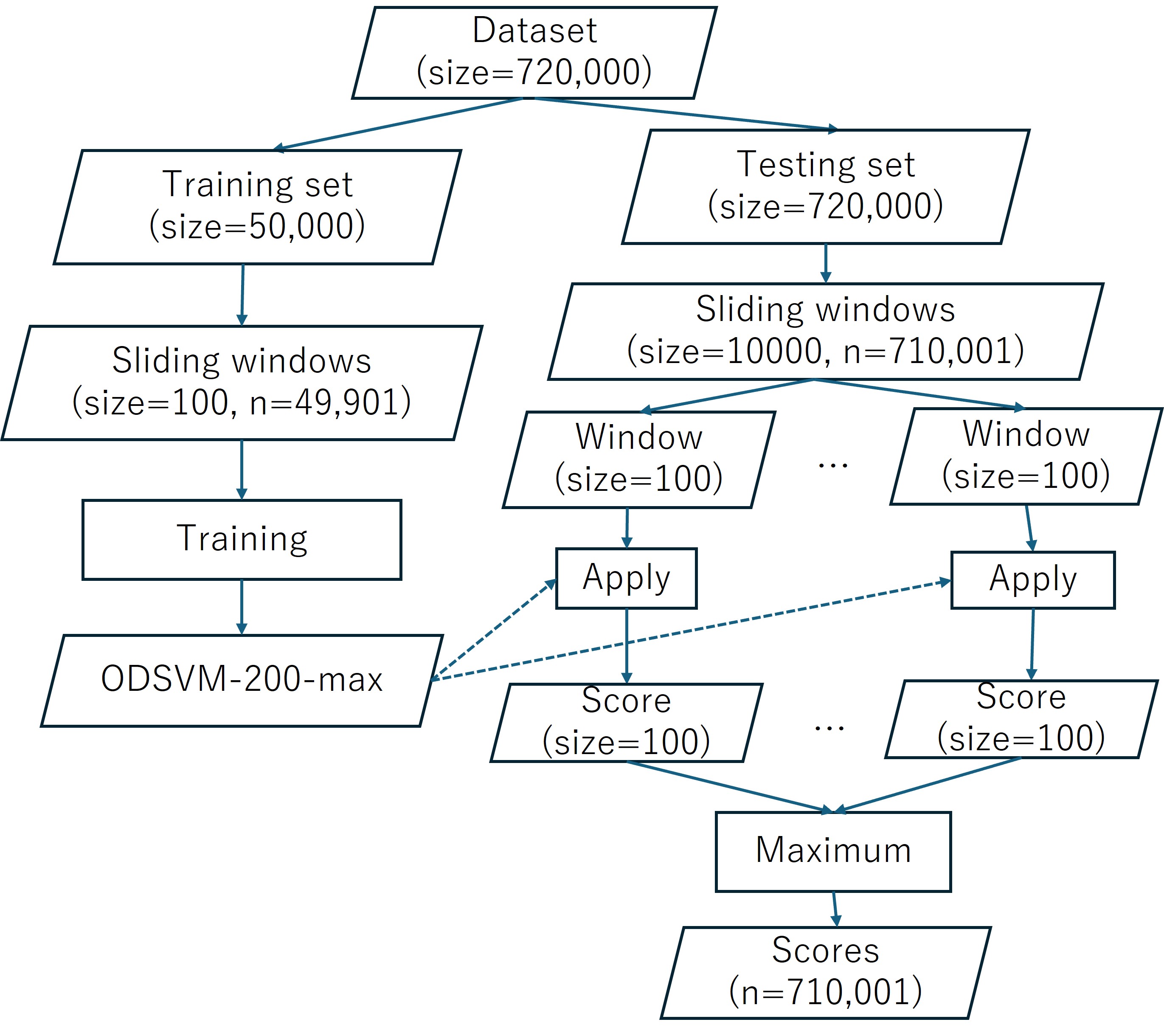}
\caption{The experiment process for breathing disorder dataset}\label{fig2}
\end{figure}

This subsection demonstrates ODSVM using a time-series dataset. The experiment uses the Ballistocardiogram (BCG) breathing disorder dataset, collected by the University of Hradec Kralove (UHK) \cite{[33]}. BCG is a signal corresponding to heart vibrations. The signals were collected from participants who performed the scheduled activities \cite{[34]}, such as holding breaths (30 seconds) and changing body positions. The measurement was performed for 720 seconds at 1000 Hz, thereby 720k time indices in total. Figure \ref{fig2} shows the experiment details. The experiment uses only one participant (ID 1) for simplicity. 

The training set is the first 50 seconds (50k time index), preprocessed into 49901 sliding windows of 0.1 second (100 indices) to train 200 ODSVM models. On the other hand, the testing set includes the entire 720 seconds (the first 50 seconds overlap with the training set), divided into 10-second sliding windows (10k indices). Subsequently, we separated these sliding windows into 0.1-second windows and applied ODSVM models to compute the normal scores. Finally, the score of a 10-second window is computed as the maximum score of its 0.1-second windows. 

\begin{figure}[H]
\centering
\includegraphics[width=1\linewidth]{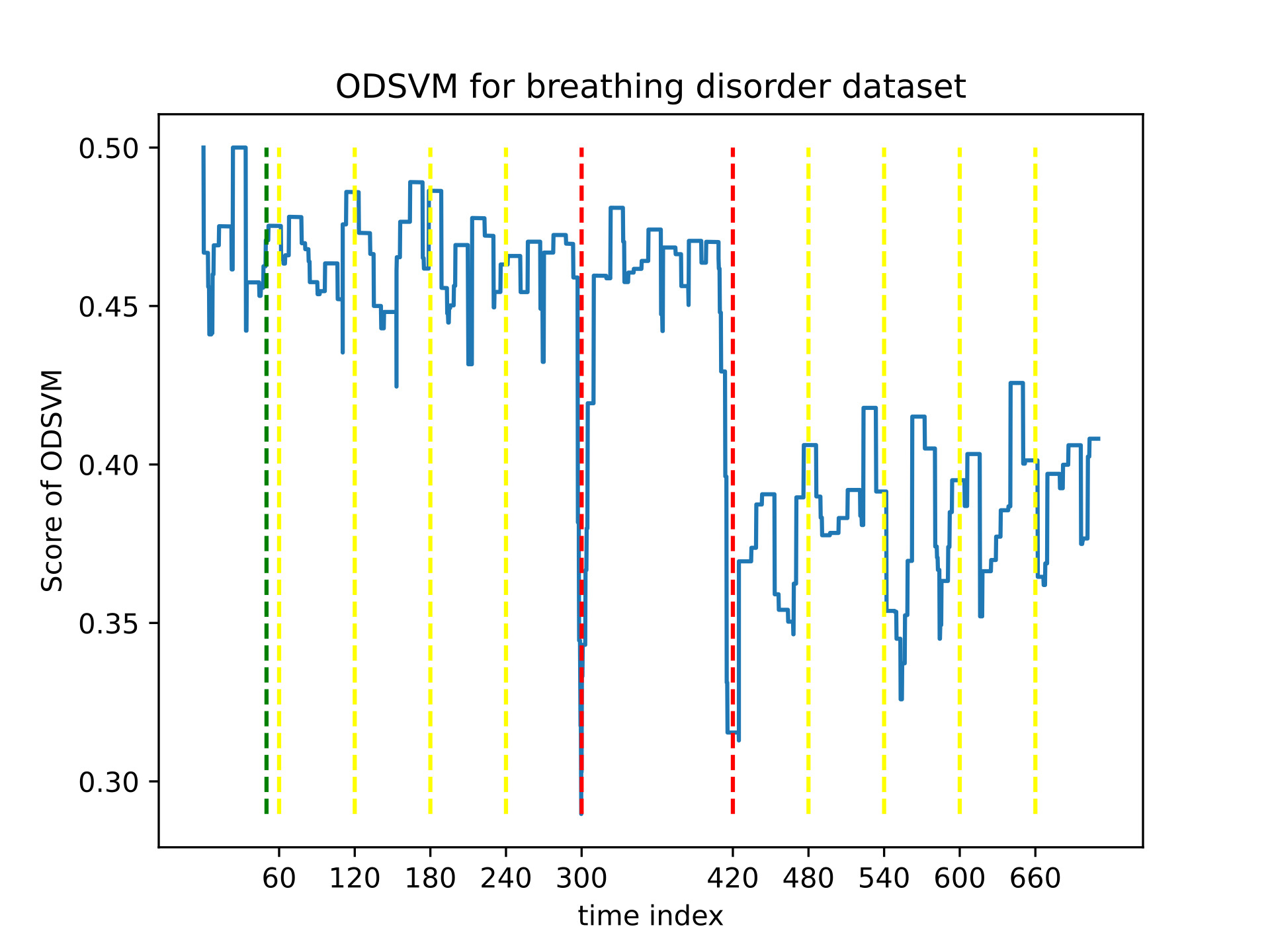}
\caption{Predicted normal scores for breathing disorder dataset}\label{fig3}
\end{figure}

Figure \ref{fig3} visualizes the normal scores for the testing set. The training set is before the green line, while the yellow and red lines show the time index of scheduled breath holds (start) and body position changes, respectively. ODSVM successfully captured anomalous vibrations caused by body position changes. Moreover, normal scores decrease while holding the breath. Furthermore, the processing took 25 seconds (the entire dataset is 720 seconds), which is a promising speed for real-time analysis of biometric signals. 

\section{Discussion}
\label{sec5}
This section discusses several aspects of ODSVM. Subsection \ref{sec5.1} shows the advantages and limitations. Subsection \ref{sec5.2} discusses the idea for decomposing other OCC algorithms. Subsection \ref{sec5.3} runs experiments with other kernels. Subsection \ref{sec5.4} reports the AUC of base learners trained on single data points. Subsection \ref{sec5.5} analyzes the number of base learners. Subsection \ref{sec5.6} compares the equations of the original OCSVM and the proposed ODSVM. Subsection \ref{sec5.7} suggests the future direction of the choice of support vectors and ensemble strategy. 

\subsection{The advantages and limitations}
\label{sec5.1}
The main advantage is processing speed because ODSVM does not solve the optimization problem. Decomposing dataset into samples can reduce time complexity from O($n^p$) to $O(n) = \sum_{i=1}^n{O(1^p)}$, where p=3 for the original, or p=2 for linear variant.

Moreover, this paper offers faster support vector selection. Support vectors are the most abnormal training samples. These samples can be determined without the optimization. Such an process could delete time complexity from 
SVM is rather outdated technique due to the rise of deep learning. SVM had several problems, such as scalability, and necessity of feature engineering.

Another advantage is one-to-one correspondence between the sample and the ODSVM model. This aspect is theoretically useful. First, representing all data as ODSVM models could protect privacy: Understanding models is more difficult than understanding data. In addition, the correspondence is beneficial for machine unlearning process. One can remove the ODSVM model corresponding to the target sample to be unlearned. This process could be simpler than traditional complex unlearning process \cite{[37]}. Moreover, all data can serve as OCC models, allowing the computation of “classification accuracy of data”. Alternatively, one can classify ODSVM models as samples.

On the other hand, AUC is sensitive to the choice of hyperparameters. Although combining all hyperparameters \cite{[21]} is a potential solution, increasing the number of base learners will reduce processing speed.

\subsection{Decomposing other OCC algorithms to one-data classification}
\label{sec5.2}
One possible extension is to train other algorithms with a single sample. However, OCC algorithms in the sklearn package cannot learn from a single sample. For instance, Local Outlier Factor \cite{[9]} needs neighbors to compute the normal score. However, single training data does not have a neighbor. Moreover, isolation Forest \cite{[8]} requires at least three samples to create a meaningful model. When the training sample size is less than 3, the model will output the same scores for all samples. In addition, the Gaussian Mixture model \cite{[10]} requires at least two training samples. On the other hand, deep learning techniques, such as autoencoder \cite{[7]}, can learn from a single sample. However, the goal of this study is to decompose OCSVM. Therefore, we leave the idea for future work.

\subsection{The experiment with other kernels}
\label{sec5.3}
One question is whether the approximation process is sensitive to hyperparameters. Accordingly, Table \ref{table8} compares OCSVM and ODSVM using other kernels (see subsection 2.1.1 for the description). The reported result is an average AUC of 27 datasets. Overall, the ODSVM ensemble reproduced results similar to those of OCSVM for other kernels.

\begin{table}[H]
\caption{The experiment with other kernels.}\label{table8}
\centering
\scalebox{0.9}{
\begin{tabular}{l l l l l l l l} \hline
One class & Kernel & OCSVM & ODSVM-all & \multicolumn{2}{l}{ODSVM-200} & \multicolumn{2}{l}{ODSVM-1000} \\ 
 &  &  & SUM & SUM & Max & SUM & Max \\ \hline
Minority & Linear  & 68.09 & 63.6 & 68.0 & 67.4 & 68.6 & 66.7 \\ 
 & Polynomial & 69.56 & 62.4 & 67.8 & 65.3 & 68.6 & 65.1 \\ 
 & RBF (default) & 76.22 & 75.2 & 73.4 & 73.3 & 75.2 & 77.4 \\ 
 & Sigmoid & 66.17 & 63.0 & 66.3 & 64.8 & 66.9 & 64.7 \\ \hline
Majority & Linear  & 47.06 & 54.7 & 43.5 & 44.9 & 50.1 & 46.8 \\ 
 & Polynomial & 50.1 & 54.8 & 43.8 & 44.7 & 50.3 & 47.2 \\ 
 & RBF (default) & 61.41 & 60.2 & 63.5 & 55.3 & 61.1 & 60.8 \\ 
 & Sigmoid & 53.51 & 54.1 & 49.3 & 47.1 & 50.8 & 47.1 \\ \hline
\end{tabular}
}
\end{table}

\subsection{AUC of base learners}
\label{sec5.4}
This subsection computes the AUC of ODSVM models that learned single data. The experiment prepared 1000 (or less) ODSVM models for each dataset. 
Table \ref{table9} reports the best, average, and the worst AUC of base learners and ensemble results for ODSVM-1000. Note that always selecting the best and the worst base learners is not realistic. Therefore, these results are marked by brackets. The ensemble of ODSVMs outperformed average ODSVMs, while the best ODSVM is better than the ensemble. 
\begin{table}[H]
\caption{AUC scores for single ODSVM models (ODSVM-1000)}\label{table9}
\centering
\scalebox{0.9}{
\begin{tabular}{l l l l l l | l l l l l} \hline
 & \multicolumn{5}{l|}{One = minority} & \multicolumn{5}{l}{One = majority} \\ 
Dataset & \multicolumn{3}{l}{ODSVMs} & \multicolumn{2}{l|}{Ensemble} & \multicolumn{3}{l}{ODSVMs} & \multicolumn{2}{l}{Ensemble} \\ 
 & Worst & Avg & Best & Sum & Max & Worst & Avg & Best & Sum & Max \\ \hline
Ecoli & 41.1 & 87.6 & 94.5 & 91.7 & 91.6 & 5.1 & 64.6 & 91.6 & 82.5 & 41.7 \\ 
optical\_digits & 59.1 & 86.2 & 97.1 & 97.5 & 98.4 & 11.6 & 37.5 & 64.6 & 22.6 & 90.9 \\ 
Satimage & 58.2 & 83.6 & 92.0 & 91.6 & 87.7 & 15.6 & 48.9 & 65.9 & 65.4 & 74.0 \\ 
pen\_digits & 33.5 & 54.1 & 69.5 & 97.2 & 99.8 & 25.5 & 45.2 & 80.8 & 45.5 & 52.8 \\ 
Abalone & 23.0 & 70.4 & 82.4 & 80.7 & 82.8 & 17.6 & 42.5 & 77.6 & 30.7 & 27.1 \\ 
sick\_euthyroid & 49.9 & 67.2 & 77.5 & 68.1 & 83.8 & 23.0 & 47.9 & 74.7 & 43.9 & 51.1 \\ 
Spectrometer & 21.4 & 48.1 & 62.4 & 52.1 & 63.4 & 39.2 & 72.4 & 86.9 & 79.8 & 78.6 \\ 
car\_eval\_34 & 62.1 & 73.2 & 84.6 & 98.9 & 91.1 & 24.4 & 51.9 & 89.1 & 96.7 & 63.7 \\ 
Isolet & 60.2 & 82.5 & 92.8 & 92.5 & 95.4 & 12.0 & 45.1 & 65.6 & 41.7 & 68.6 \\ 
us\_crime & 18.6 & 62.9 & 89.1 & 69.6 & 73.0 & 17.4 & 69.1 & 88.9 & 80.1 & 76.1 \\ 
yeast\_ml8 & 40.7 & 52.9 & 64.4 & 58.4 & 51.4 & 35.4 & 49.4 & 64.8 & 51.9 & 57.5 \\ 
Scene & 43.2 & 59.3 & 73.3 & 61.5 & 62.6 & 27.2 & 48.8 & 68.7 & 48.7 & 46.1 \\ 
libras\_move & 37.6 & 61.4 & 83.1 & 89.7 & 96.2 & 28.7 & 55.1 & 82.2 & 73.8 & 92.1 \\ 
thyroid\_sick & 39.5 & 70.0 & 80.8 & 73.0 & 78.3 & 21.1 & 55.3 & 78.4 & 55.8 & 56.6 \\ 
coil\_2000 & 36.3 & 52.8 & 65.1 & 54.6 & 58.5 & 34.7 & 49.1 & 64.5 & 47.7 & 46.8 \\ 
Arrhythmia & 52.0 & 59.3 & 68.7 & 61.0 & 67.6 & 27.1 & 43.1 & 60.8 & 43.4 & 44.3 \\
solar\_flare\_m0 & 23.4 & 51.9 & 81.6 & 52.7 & 52.4 & 18.9 & 64.7 & 82.5 & 78.1 & 67.5 \\
Oil & 40.2 & 54.1 & 72.3 & 57.6 & 68.5 & 26.3 & 53.2 & 67.5 & 64.3 & 85.8 \\
car\_eval\_4 & 74.4 & 81.9 & 89.7 & 99.6 & 94.2 & 14.8 & 51.3 & 92.1 & 97.1 & 55.9 \\ 
wine\_quality & 31.9 & 48.6 & 68.3 & 48.4 & 59.4 & 33.9 & 57.4 & 71.5 & 68.2 & 61.2 \\
letter\_img & 50.2 & 76.1 & 90.6 & 88.4 & 99.3 & 19.7 & 59.0 & 84.1 & 78.0 & 65.1 \\
yeast\_me2 & 23.3 & 70.5 & 87.5 & 78.5 & 84.4 & 12.1 & 65.8 & 87.7 & 75.9 & 62.0 \\
Webpage & 27.9 & 59.8 & 69.1 & 61.7 & 87.7 & 29.4 & 66.9 & 81.2 & 76.4 & 89.1 \\
ozone\_level & 24.3 & 72.6 & 83.3 & 79.9 & 81.5 & 17.9 & 52.2 & 84.4 & 40.9 & 24.3 \\
Mammography & 13.4 & 72.8 & 89.3 & 80.0 & 47.9 & 15.7 & 41.6 & 86.7 & 37.4 & 28.2 \\
protein\_homo & 15.7 & 64.3 & 94.7 & 78.0 & 70.9 & 4.7 & 63.1 & 94.3 & 64.3 & 68.8 \\
Abalone\_19 & 38.2 & 62.8 & 69.5 & 68.0 & 63.2 & 29.8 & 51.0 & 69.2 & 60.1 & 64.4 \\
Avg & (38.5) & 66.2 & (80.5) & 75.2 & 77.4 & (21.8) & 53.7 & (78.0) & 61.1 & 60.8 \\ \hline

\end{tabular}}
\end{table}

It is interesting to note that ODSVM models learned only single data points; the model from a single data outperformed the model from the dataset. However, determining the best ODSVM model is challenging without accessing testing samples. This feature could be useful in decomposing binary or multi-class classification into OCC. 

\subsection{Number of base learners}
\label{sec5.5}
This subsection analyzes the relation between the number of ODSVMs and the AUC of ensemble models. For this purpose, the experiment trains 1000 (or fewer) ODSVM models and combines these models incrementally. The combination begins with the model trained on the most abnormal sample.

\begin{figure}[H]
\centering
\includegraphics[width=1\linewidth]{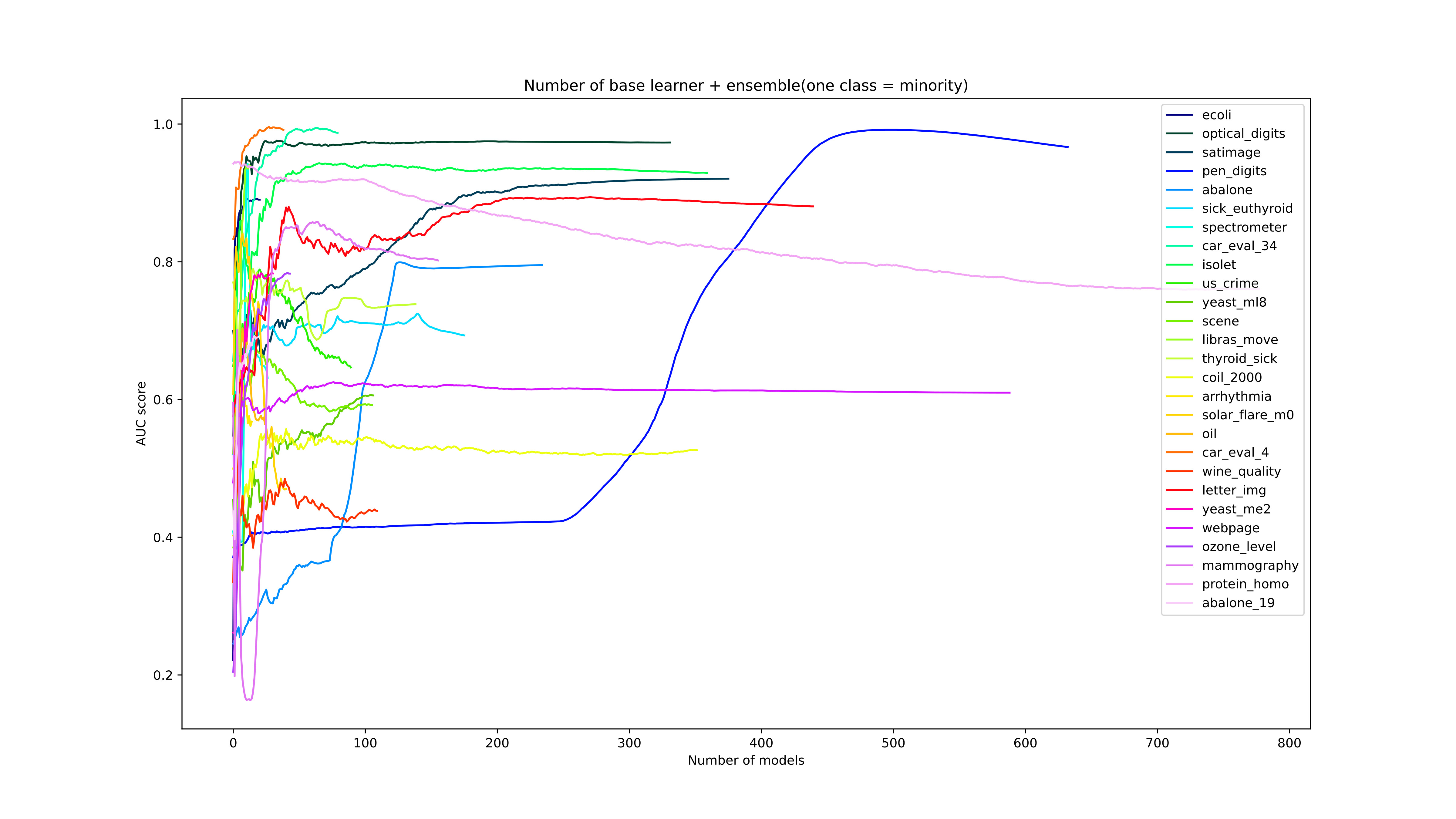}
\caption{The relation between number of models and AUC (One class is minority)}\label{fig4}
\end{figure}

Figure \ref{fig4} reports the AUCs for 27 datasets in which one class is a minority. The reported result is ODSVM-N SUM, where N is the number of base learners. The lines disappear when N is larger than the number of training samples. Generally, AUC increases with more base learners. In particular, the pen\_digits dataset (blue line) shows 40\%  AUC until 250 base learners, and the result suddenly improved around 300 models. 

\begin{figure}[H]
\centering
\includegraphics[width=1\linewidth]{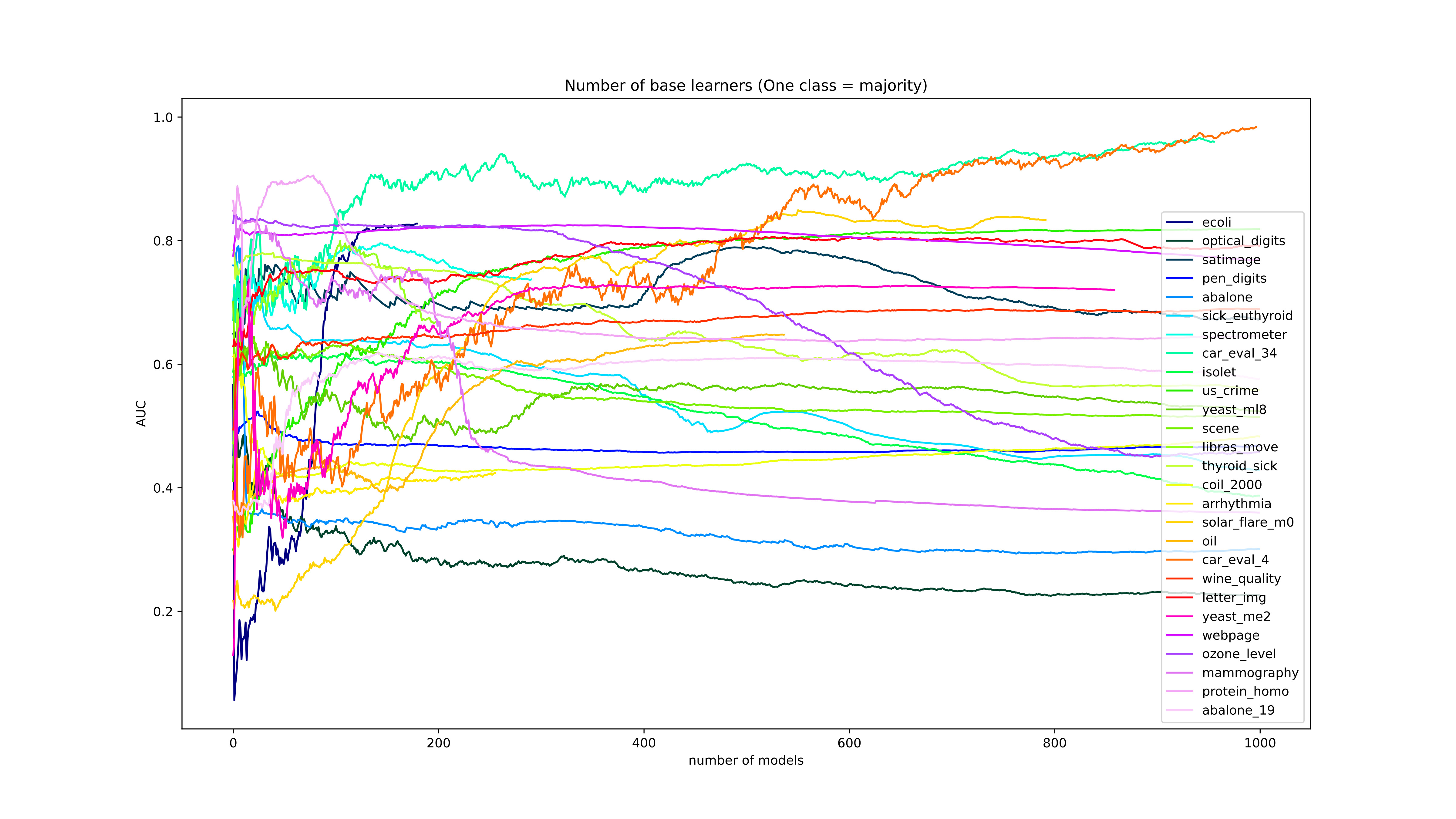}
\caption{The relation between number of models and AUC (One class is majority)}\label{fig5}
\end{figure}

Figure \ref{fig5} shows the same result when one class is the majority. Car\_eval\_34(green line) and Car\_eval\_4 (orange line) improve AUC as the models increase, whereas most datasets degrade AUC.

\begin{figure}[H]
\centering
\includegraphics[width=1\linewidth]{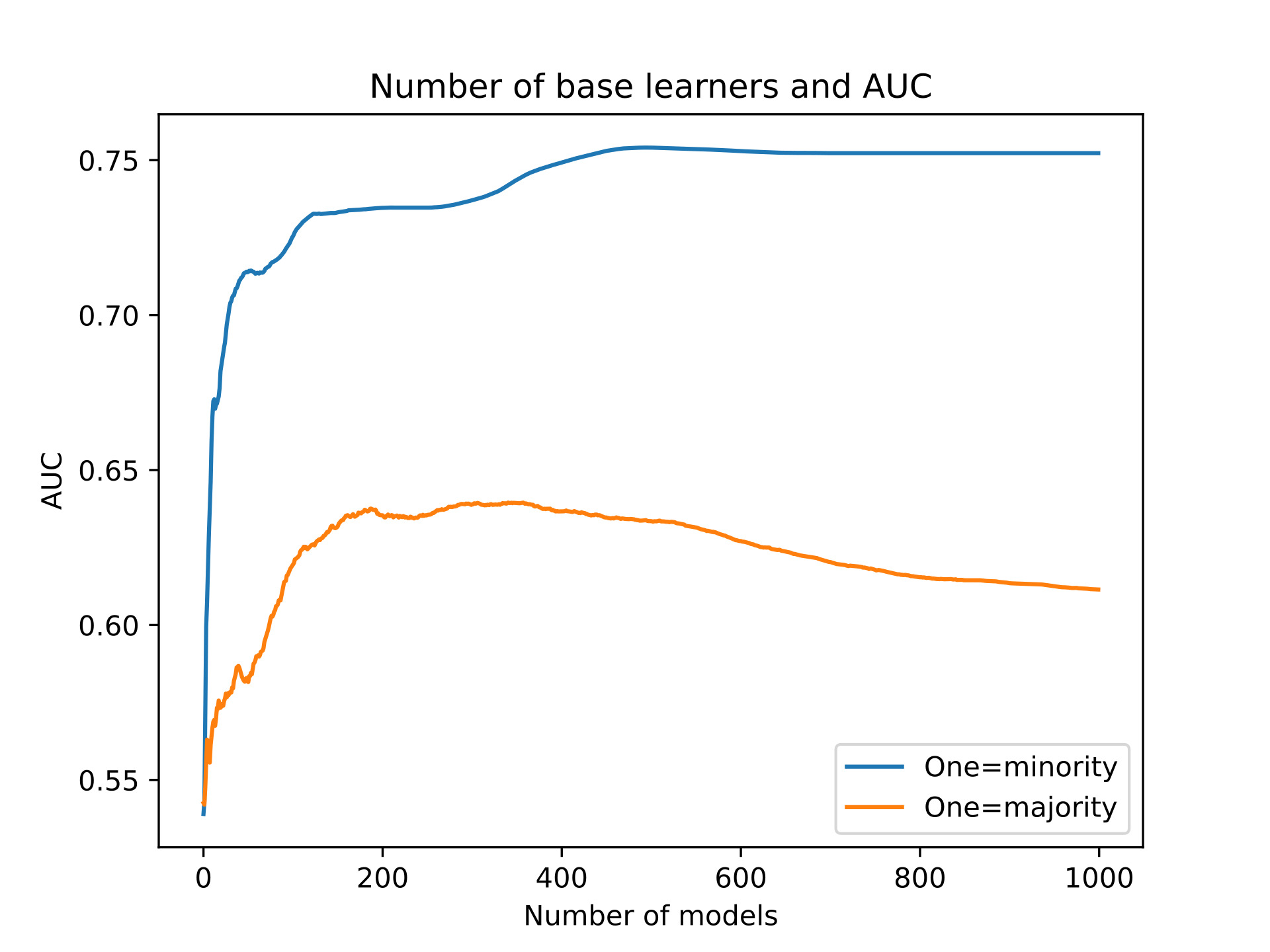}
\caption{Number of base learners and AUC (average of all datasets)}\label{fig6}
\end{figure}

Figure \ref{fig6} shows average AUC scores for ensemble results. This is a simplified version of Figure \ref{fig4} and Figure \ref{fig5}. Minority class improves with the number of models, while majority class decreases AUC.

\subsection{The comparison of equations}
\label{sec5.6}
Table \ref{table10} compares the optimization problems and score functions between traditional OCSVM and ODSVM. The traditional OCSVM optimized $n^2$ weights, which were time-consuming. On the other hand, the proposed method can decompose the dual problem into N obvious problems. 

Regarding the score function, traditional OCSVM \cite{[2]} computed individual weights $\alpha_i$ by optimization. On the other hand, ODSVM-N-sum provided the same weight for all support vectors, while ODSVM-N-max assigns the maximum score, giving all weight to the nearest support vector. All experiment results showed the similar AUC scores. 

Support vectors are samples around borderline between classes. In OCSVM (and other SVM variants), only support vectors are important samples. Traditional OCSVM considered optimization to determine support vectors. However, this process was unnecessarily complex and had to include other useless samples. Support vectors of OCSVM are the most abnormal training samples. One can determine abnormal samples without the optimization.

On the other hand, the weight of support vectors are still debatable topic. One reasonable hypothesis is that similar support vectors are not necessary. So their weights could be controlled based on support vectors of similarity. 

Perhaps, similar support vectors are useless; one sample is enough. So the weights could be determined based on the similarity between support vectors. This optimization could be faster than including unimportant samples. 

\begin{table}[H]
\caption{Comparison of equations}\label{table10}
\centering
\scalebox{1}{
\begin{tabular}{l l l} \hline
Method & Optimization problem & Score function \\ \hline
\\
OCSVM & $\min_{\alpha}⁡ \frac{1}{2}\sum_{i=1}^{n}\sum_{j=1}^{n}{\alpha_i}{\alpha_j}K(x_i,x)$ & $\sum_{i=1}^n \alpha_i K(x_i, x)-\rho$ \\ \hline
\\
ODSVM-N-sum & \multirow{2}{*}{$\frac{1}{2} K(x_{tr},x_{tr})$ * N times} & $\sum_{i=1}^N (K(x_i, x)-\rho)$ \\
ODSVM-N-max & & $\max_i{⁡(K(x_i,x)}-\rho)$ \\ \hline

\end{tabular}
}
\end{table}

\subsection{Choice of support vectors and ensemble strategy}
\label{sec5.7}
The proposed ODSVM successfully accelerated OCSVM. Decomposing the dataset into samples is a promising direction. However, support vector selection and ensemble strategies need further discussion. Algorithm \ref{alg5} suggests future work to consider an ideal ODSVM framework. There are three important questions: 1) how many support vectors are required, 2) how to select support vectors; the process is unsupervised outlier detection, and 3) how to combine ODSVM models trained from selected support vectors. Moreover, all process must be executed fast. The core heuristic is that support vectors are the most abnormal samples.

\begin{algorithm}
\caption{Ideal ODVM framework (Future work)}\label{alg5}
1.	Select \textbf{an ideal number} of support vectors.

2.	\textbf{Apply an ideal strategy to select support vectors} from dataset.

3.	Train ODSVM models from support vectors.

4.	Apply \textbf{an ideal ensemble strategy} to combine ODSVM models.
\end{algorithm}

\section{Conclusion and future work}
\label{sec6}
This paper decomposed OCSVM into an ensemble of ODSVM models trained on single data points. Moreover, the proposal introduces simple heuristic to treat most abnormal training samples as support vectors. These ideas accelerated OCSVM significantly by deleting large scale optimization and support vector selection.

Future work should consider the details of support vector selection and the ensemble strategy for improving AUC scores. Moreover, the idea could be extended to decompose deep learning to a single data level. In addition, training the model on a single data point could contribute to the classification of the machine learning models. 

\section*{CRediT authorship contribution statement}
\textbf{Toshitaka Hayashi}: Conceptualization, Methodology, Writing – original draft, Investigation, Software, Visualization. \textbf{Dalibor Cimr}: Writing – original draft \textbf{Hamido Fujita}: Writing – review and editing, Supervision. \textbf{Richard Cimler}: Project administration, Funding acquisition. 

\section*{Declaration of Competing Interest}
\label{decCI}
The authors declare that they have no known competing financial interests or personal relationships that could have appeared to influence the work reported in this paper.

\section*{Acknowledgement}
\label{ack}
This study is supported by the research project “2200/04/2024-2026” as part of the "Competition for 2024-2026 Postdoctoral Job Positions at the University of Hradec Králové", at the Faculty of Science, University of Hradec Králové. This study is supported from the project "Research of Excellence on Digital Technologies and Wellbeing CZ.02.01.01/00/22\_008/0004583“, which is co-financed by the European Union.


\bibliography{reference.bib}

@article{[1],
  title={Estimating the support of a high-dimensional distribution},
  author={Sch{\"o}lkopf, Bernhard and Platt, John C and Shawe-Taylor, John and Smola, Alex J and Williamson, Robert C},
  journal={Neural computation},
  volume={13},
  number={7},
  pages={1443--1471},
  year={2001},
  publisher={MIT Press}
}

@article{[2],
  title={Clustering-based ensembles for one-class classification},
  author={Krawczyk, Bartosz and Wo{\'z}niak, Micha{\l} and Cyganek, Bogus{\l}aw},
  journal={Information sciences},
  volume={264},
  pages={182--195},
  year={2014},
  publisher={Elsevier}
}

@article{[3],
  title={On the usefulness of one-class classifier ensembles for decomposition of multi-class problems},
  author={Krawczyk, Bartosz and Wo{\'z}niak, Micha{\l} and Herrera, Francisco},
  journal={Pattern Recognition},
  volume={48},
  number={12},
  pages={3969--3982},
  year={2015},
  publisher={Elsevier}
}

@article{[4],
  title={Approximate training of one-class support vector machines using expected margin},
  author={Kang, Seokho and Kim, Dongil and Cho, Sungzoon},
  journal={Computers \& Industrial Engineering},
  volume={130},
  pages={772--778},
  year={2019},
  publisher={Elsevier}
}

@article{[5],
  title={One-class support vector classifiers: A survey},
  author={Alam, Shamshe and Sonbhadra, Sanjay Kumar and Agarwal, Sonali and Nagabhushan, P},
  journal={Knowledge-Based Systems},
  volume={196},
  pages={105754},
  year={2020},
  publisher={Elsevier}
}

@article{[6],
  title={Scikit-learn: Machine learning in Python},
  author={Pedregosa, Fabian and Varoquaux, Ga{\"e}l and Gramfort, Alexandre and Michel, Vincent and Thirion, Bertrand and Grisel, Olivier and Blondel, Mathieu and Prettenhofer, Peter and Weiss, Ron and Dubourg, Vincent and others},
  journal={the Journal of machine Learning research},
  volume={12},
  pages={2825--2830},
  year={2011},
  publisher={JMLR. org}
}

@article{[7],
  title={Reducing the dimensionality of data with neural networks},
  author={Hinton, Geoffrey E and Salakhutdinov, Ruslan R},
  journal={science},
  volume={313},
  number={5786},
  pages={504--507},
  year={2006},
  publisher={American Association for the Advancement of Science}
}

@inproceedings{[8],
  title={Isolation forest},
  author={Liu, Fei Tony and Ting, Kai Ming and Zhou, Zhi-Hua},
  booktitle={2008 eighth ieee international conference on data mining},
  pages={413--422},
  year={2008},
  organization={IEEE}
}

@inproceedings{[9],
  title={LOF: identifying density-based local outliers},
  author={Breunig, Markus M and Kriegel, Hans-Peter and Ng, Raymond T and Sander, J{\"o}rg},
  booktitle={Proceedings of the 2000 ACM SIGMOD international conference on Management of data},
  pages={93--104},
  year={2000}
}

@article{[10],
  title={Unsupervised learning of finite mixture models},
  author={Figueiredo, Mario A. T. and Jain, Anil K.},
  journal={IEEE Transactions on pattern analysis and machine intelligence},
  volume={24},
  number={3},
  pages={381--396},
  year={2002},
  publisher={Ieee}
}

@article{[11],
  title={Average Localised Proximity: A new data descriptor with good default one-class classification performance},
  author={Lenz, Oliver Urs and Peralta, Daniel and Cornelis, Chris},
  journal={Pattern Recognition},
  volume={118},
  pages={107991},
  year={2021},
  publisher={Elsevier}
}

@article{[12],
  title={Support vector data description},
  author={Tax, David MJ and Duin, Robert PW},
  journal={Machine learning},
  volume={54},
  number={1},
  pages={45--66},
  year={2004},
  publisher={Springer}
}

@inproceedings{[13],
  title={Subspace support vector data description},
  author={Sohrab, Fahad and Raitoharju, Jenni and Gabbouj, Moncef and Iosifidis, Alexandros},
  booktitle={2018 24th International Conference on Pattern Recognition (ICPR)},
  pages={722--727},
  year={2018},
  organization={IEEE}
}

@article{[14],
  title={Imbalanced-learn: A python toolbox to tackle the curse of imbalanced datasets in machine learning},
  author={Lema{\~A}{\v{Z}}tre, Guillaume and Nogueira, Fernando and Aridas, Christos K},
  journal={Journal of machine learning research},
  volume={18},
  number={17},
  pages={1--5},
  year={2017}
}

@article{[15],
  title={Least squares one-class support vector machine},
  author={Choi, Young-Sik},
  journal={Pattern Recognition Letters},
  volume={30},
  number={13},
  pages={1236--1240},
  year={2009},
  publisher={Elsevier}
}

@article{[16],
  title={Robust AdaBoost based ensemble of one-class support vector machines},
  author={Xing, Hong-Jie and Liu, Wei-Tao},
  journal={Information Fusion},
  volume={55},
  pages={45--58},
  year={2020},
  publisher={Elsevier}
}

@article{[18],
  title={One-class SVMs for document classification},
  author={Manevitz, Larry M and Yousef, Malik},
  journal={Journal of machine Learning research},
  volume={2},
  number={Dec},
  pages={139--154},
  year={2001}
}

@inproceedings{[19],
  title={Improving one-class SVM for anomaly detection},
  author={Li, Kun-Lun and Huang, Hou-Kuan and Tian, Sheng-Feng and Xu, Wei},
  booktitle={Proceedings of the 2003 international conference on machine learning and cybernetics (IEEE Cat. No. 03EX693)},
  volume={5},
  pages={3077--3081},
  year={2003},
  organization={IEEE}
}

@article{[20],
  title={Support-vector networks},
  author={Cortes, Corinna and Vapnik, Vladimir},
  journal={Machine learning},
  volume={20},
  number={3},
  pages={273--297},
  year={1995},
  publisher={Springer}
}

@article{[21],
  title={The fusion of hyperparameter candidates for one-class classification problems},
  author={Hayashi, Toshitaka and Cimr, Dalibor and Fujita, Hamido and Cimler, Richard and Aljuaid, Hanan},
  journal={Information Sciences},
  pages={122526},
  year={2025},
  publisher={Elsevier}
}

@article{[22],
  title={Sample reduction using farthest boundary point estimation (FBPE) for support vector data description (SVDD)},
  author={Alam, Shamshe and Sonbhadra, Sanjay Kumar and Agarwal, Sonali and Nagabhushan, P and Tanveer, Muhammad},
  journal={Pattern Recognition Letters},
  volume={131},
  pages={268--276},
  year={2020},
  publisher={Elsevier}
}

@article{[24],
  title={Fuzzy one-class support vector machines},
  author={Hao, Pei-Yi},
  journal={Fuzzy Sets and Systems},
  volume={159},
  number={18},
  pages={2317--2336},
  year={2008},
  publisher={Elsevier}
}

@article{[25],
  title={Brain activation detection by neighborhood one-class SVM},
  author={Yang, Jian and Zhong, Ning and Liang, Peipeng and Wang, Jue and Yao, Yiyu and Lu, Shengfu},
  journal={Cognitive Systems Research},
  volume={11},
  number={1},
  pages={16--24},
  year={2010},
  publisher={Elsevier}
}

@article{[26],
  title={A survey of ensemble learning: Concepts, algorithms, applications, and prospects},
  author={Mienye, Ibomoiye Domor and Sun, Yanxia},
  journal={Ieee Access},
  volume={10},
  pages={99129--99149},
  year={2022},
  publisher={IEEE}
}

@article{[27],
  title={One-class convolutional neural network},
  author={Oza, Poojan and Patel, Vishal M},
  journal={IEEE Signal Processing Letters},
  volume={26},
  number={2},
  pages={277--281},
  year={2018},
  publisher={IEEE}
}

@article{[28],
  title={Deep anomaly detection using geometric transformations},
  author={Golan, Izhak and El-Yaniv, Ran},
  journal={Advances in neural information processing systems},
  volume={31},
  year={2018}
}

@article{[29],
  title={Bagging predictors},
  author={Breiman, Leo},
  journal={Machine learning},
  volume={24},
  number={2},
  pages={123--140},
  year={1996},
  publisher={Springer}
}

@article{[30],
  title={Ensemble based systems in decision making},
  author={Polikar, Robi},
  journal={IEEE Circuits and systems magazine},
  volume={6},
  number={3},
  pages={21--45},
  year={2006},
  publisher={IEEE}
}

@article{[31],
  title={Stacked generalization},
  author={Wolpert, David H},
  journal={Neural networks},
  volume={5},
  number={2},
  pages={241--259},
  year={1992},
  publisher={Elsevier}
}

@article{[32],
  title={A decision-theoretic generalization of on-line learning and an application to boosting},
  author={Freund, Yoav and Schapire, Robert E},
  journal={Journal of computer and system sciences},
  volume={55},
  number={1},
  pages={119--139},
  year={1997},
  publisher={Elsevier}
}

@misc{[33],
author = {Studnicka, Filip},
title = {Ballistocardiography with breathing disorderes},
doi = {10.17632/9fmfn6kfn7.3},
howpublished= {Mendeley data},
year={2022}}

@article{[34],
  title={Computer aided detection of breathing disorder from ballistocardiography signal using convolutional neural network},
  author={Cimr, Dalibor and Studnicka, Filip and Fujita, Hamido and Tomaskova, Hana and Cimler, Richard and Kuhnova, Jitka and Slegr, Jan},
  journal={Information Sciences},
  volume={541},
  pages={207--217},
  year={2020},
  publisher={Elsevier}
}

@article{[35],
  title={One-class classification: taxonomy of study and review of techniques},
  author={Khan, Shehroz S and Madden, Michael G},
  journal={The Knowledge Engineering Review},
  volume={29},
  number={3},
  pages={345--374},
  year={2014},
  publisher={Cambridge University Press}
}

@inproceedings{[36],
  title={Deep one-class classification},
  author={Ruff, Lukas and Vandermeulen, Robert and Goernitz, Nico and Deecke, Lucas and Siddiqui, Shoaib Ahmed and Binder, Alexander and M{\"u}ller, Emmanuel and Kloft, Marius},
  booktitle={International conference on machine learning},
  pages={4393--4402},
  year={2018},
  organization={PMLR}
}

@article{[37],
  title={A survey of machine unlearning},
  author={Nguyen, Thanh Tam and Huynh, Thanh Trung and Ren, Zhao and Nguyen, Phi Le and Liew, Alan Wee-Chung and Yin, Hongzhi and Nguyen, Quoc Viet Hung},
  journal={ACM Transactions on Intelligent Systems and Technology},
  volume={16},
  number={5},
  pages={1--46},
  year={2025},
  publisher={ACM New York, NY}
}

@inproceedings{subsetSVM,
  title={ENSEMBLE RANDOM-SUBSET SVM},
  author={Nishida, Kenji and Fujiki, Jun and Kurita, Takio},
  booktitle={International Conference on Neural Computation Theory and Applications},
  volume={2},
  pages={334--339},
  year={2011},
  organization={SCITEPRESS}
}

@article{NystromSVM,
  title={Using the Nystr{\"o}m method to speed up kernel machines},
  author={Williams, Christopher and Seeger, Matthias},
  journal={Advances in neural information processing systems},
  volume={13},
  year={2000}
}

@article{KDE,
  title={Remarks on Some Nonparametric Estimates of a Density Function},
  author={Murray Rosenblatt},
  journal={Annals of Mathematical Statistics},
  year={1956},
  volume={27},
  pages={832-837},
}
\bibliographystyle{elsarticle-num} 

\end{document}